\documentclass[letterpaper, 10 pt, conference]{ieeeconf}  % Comment this line out if you need a4paper

\IEEEoverridecommandlockouts                              % This command is only needed if 
\title{\LARGE \bf
Enhancing Situational Awareness in Underwater Robotics with Multi-modal Spatial Perception
}

\author{Pushyami Kaveti$^{1}$, Ambjørn Grimsrud Waldum$^{2}$, Hanumant Singh$^{1}$, and Martin Ludvigsen$^{2}$% <-this % stops a space
\thanks{$^{1}$P. Kaveti, H. Singh, are with Northeastern University, USA {\tt\scriptsize \{kaveti.p, ha. singh\} @northeastern.edu}}. \thanks{$^{2}$ Ambjørn Waldum and Martin Ludvigsen are with NTNU, Norway {\tt\scriptsize \{ambjorn.waldum, martin.ludvigsen\}@ntnu.no}}%
}
\usepackage{graphics} % for pdf, bitmapped graphics files
\usepackage{epsfig} % for postscript graphics files
\usepackage{mathptmx} % assumes new font selection scheme installed
\usepackage{times} % assumes new font selection scheme installed
\usepackage{amsmath} % assumes amsmath package installed
\usepackage{amssymb}  % assumes amsmath package installed
\usepackage{subcaption}
\usepackage{makecell}
\usepackage{soul}
\usepackage{url}
\usepackage{siunitx}
\usepackage{graphicx}
\usepackage{mwe} 
\usepackage{booktabs}
\usepackage[table]{xcolor}
\usepackage{multirow}
\usepackage{cleveref}
\usepackage{fancyhdr}
\fancypagestyle{withfooter}{
  
  \fancyhead[L]{}
  \fancyhead[R]{}
  \fancyfoot[C]{\footnotesize Presented at the 2025 IEEE ICRA Workshop on Field Robotics}
}
\begin{document}

\maketitle
\thispagestyle{withfooter}
\pagestyle{withfooter}

%%%%%%%%%%%%%%%%%%%%%%%%%%%%%%%%%%%%%%%%%%%%%%%%%%%%%%%%%%%%%%%%%%%%%%%%%%%%%%%%
\begin{abstract}
Autonomous Underwater Vehicles (AUVs) and Remotely Operated Vehicles (ROVs) demand robust spatial perception capabilities, including Simultaneous Localization and Mapping (SLAM), to support both remote and autonomous tasks. Vision-based systems have been integral to these advancements, capturing rich color and texture at low cost while enabling semantic scene understanding. However, underwater conditions—such as light attenuation, backscatter, and low contrast—often degrade image quality to the point where traditional vision-based SLAM pipelines fail. Moreover, these pipelines typically rely on monocular or stereo inputs, limiting their scalability to the multi-camera configurations common on many vehicles.
To address these issues, we propose to leverage multi-modal sensing that fuses data from multiple sensors—including cameras, inertial measurement units (IMUs), and acoustic devices—to enhance situational awareness and enable robust, real-time SLAM. We explore both geometric and learning-based techniques along with semantic analysis, and conduct experiments on the data collected from a work-class ROV during several field deployments in the Trondheim Fjord. Through our experimental results, we demonstrate the feasibility of real-time reliable state estimation and high-quality 3D reconstructions in visually challenging underwater conditions. We also discuss system constraints and identify open research questions, such as sensor calibration, limitations with learning-based methods, that merit further exploration to advance large-scale underwater operations.
\end{abstract}

\section{Introduction} 

Technological advances in robotics, particularly Autonomous Underwater Vehicles (AUVs) and Remotely Operated Vehicles (ROVs), have revolutionized deep ocean exploration, enabling access to remote, hazardous, and otherwise inaccessible environments \cite{ice}\cite{deep_sea_exploration}. These platforms enable many critical applications such as climate research, scientific drilling, environmental monitering, inspection \& maintenance of subsea infrastructure \cite{infrastructure_inspection}\cite{push_for_autonomy}\cite{shipwreck_inspection}. Despite their capabilities, AUVs and ROVs operating in extreme underwater conditions often face challenging issues including sensor failures, communication blackouts, noisy or degraded sensor data, and highly dynamic environmental factors. Underwater vehicles must possess a certain level of onboard intelligence and situational awareness to make informed decisions in real time to handle these difficulties. Robust and real-time spatial perception is central to achieving situational awareness in complex and unstructured environments.

\begin{figure}[h]
    \centering
    \includegraphics[width=0.6\linewidth]{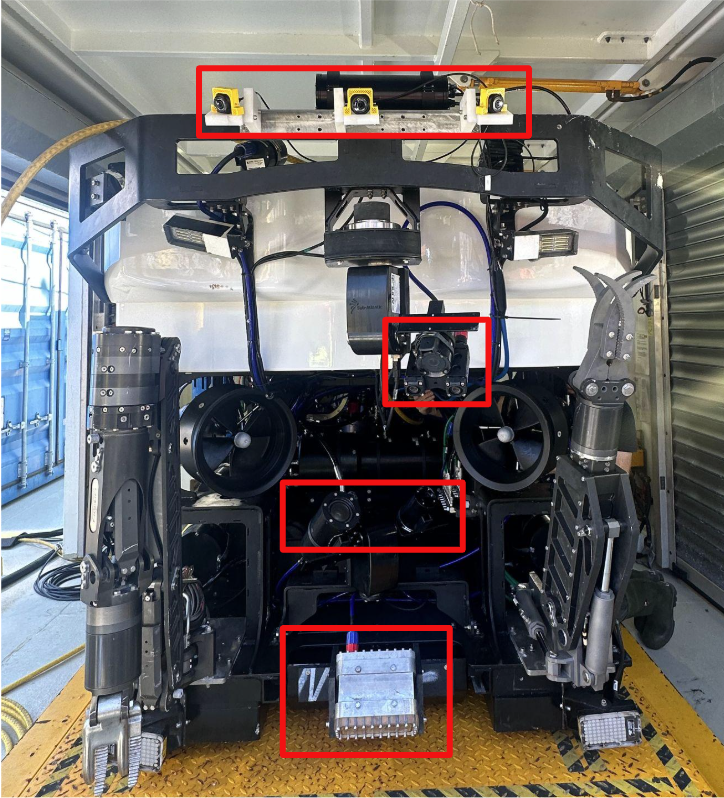}   
    \caption{Minerva II. The red-boxes from top to bottom shows the location of; Multi-camera rig. Upper HD pilot camera on its own pan-tilt unit. Lower HD pilot camera and a 4K pilot camera on the same pan-tilt unit. A stereo-camera housing.}
    \label{fig:minerva2}
\end{figure}

\begin{figure}[h]
    \centering
    \includegraphics[width=0.8\linewidth]{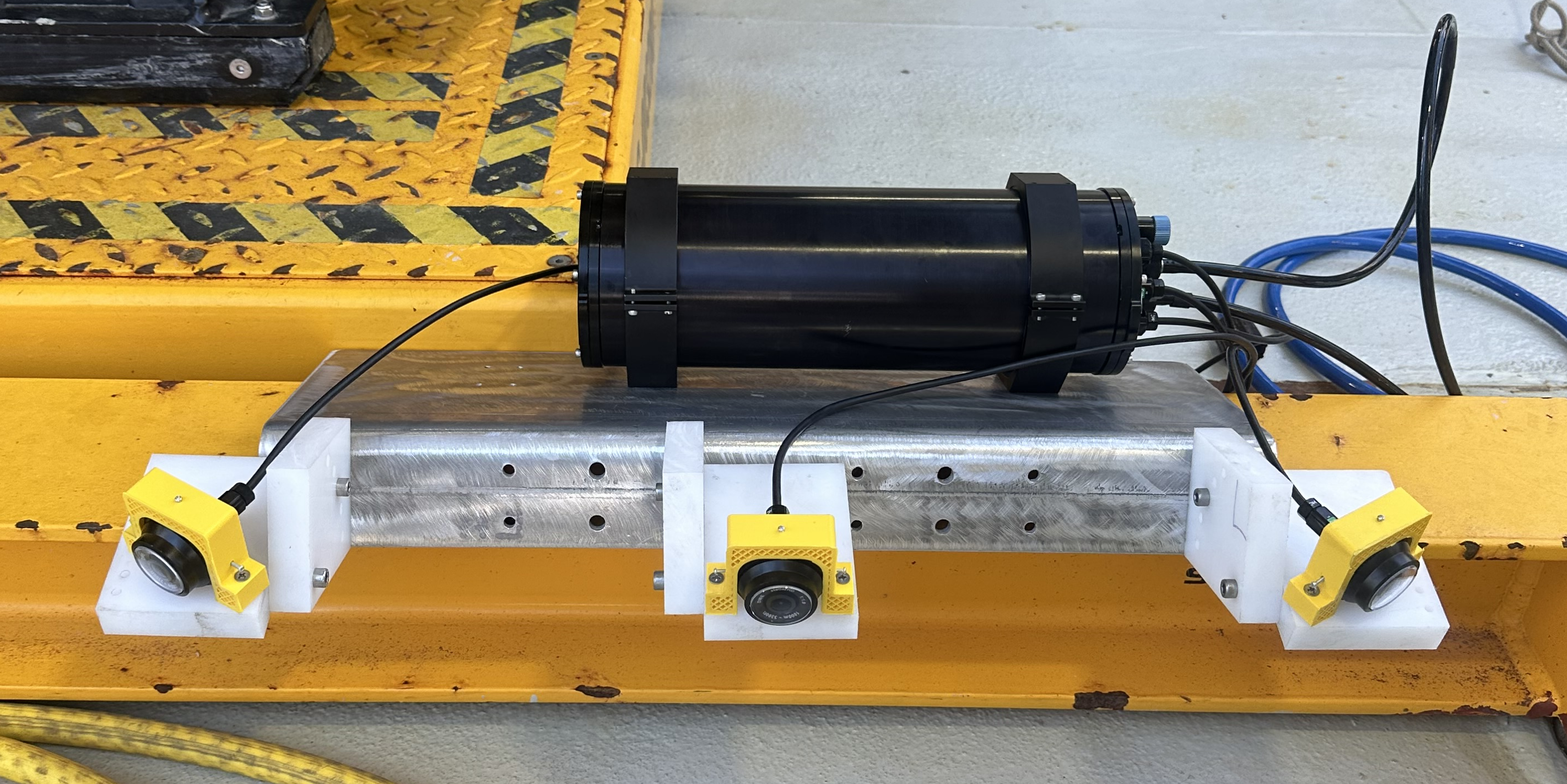}
    \caption{Prototype of multi-camera rig consisting of three Deep Water Explorer stellarHD cameras and a Blue Robotics junction bottle.}
    \label{fig:multi_camera_rig}
\end{figure}

\begin{figure}[h]
    \centering
    \includegraphics[width=0.8\linewidth]{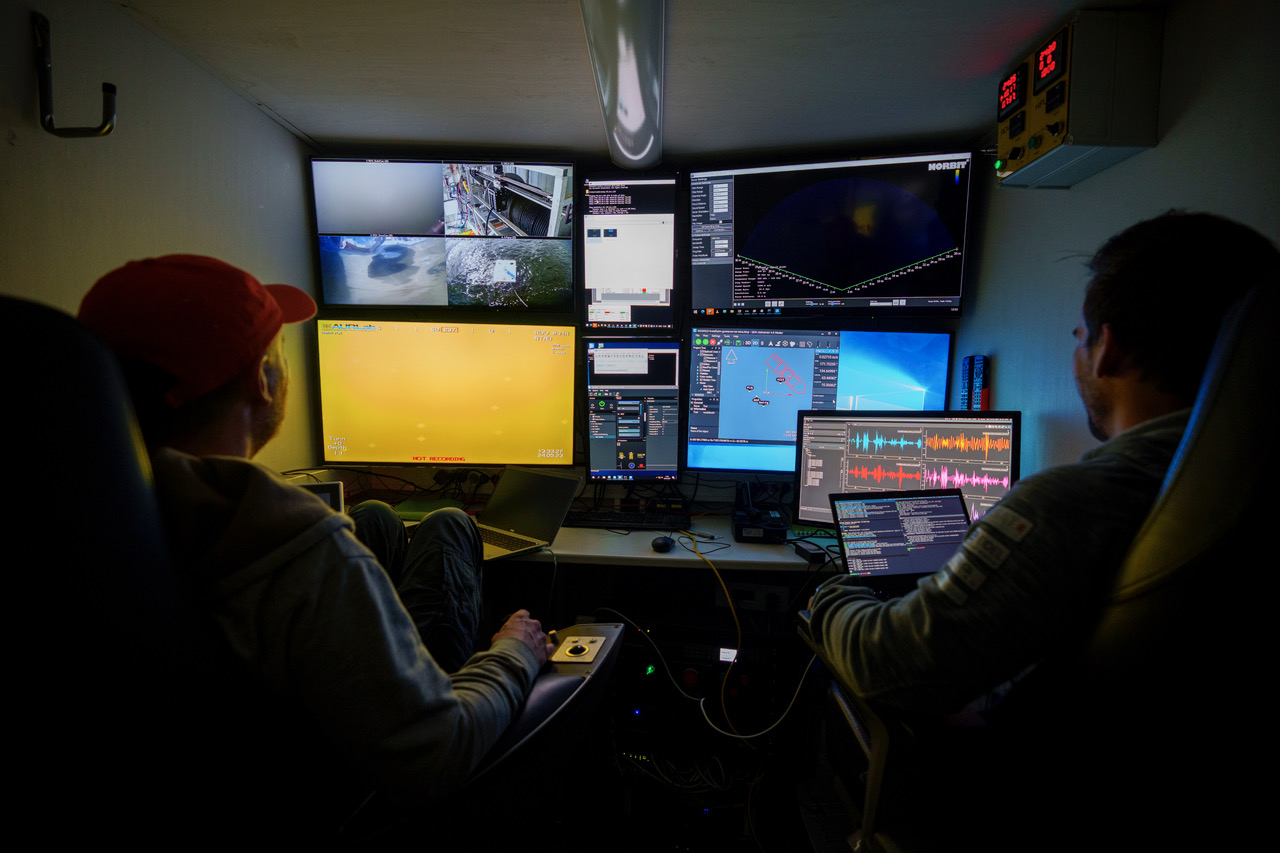}
    \caption{ROV Control Room. Courtesy of Ole Martin Wold.}
    \label{fig:minerva2_control_room}
\end{figure}

%\begin{figure}[t]
%\centering
%\begin{tabular}{c c}
%\multirow{2}{*}{ \subfloat[Minerva II. The red-boxes from top to bottom shows the %location of; Our Multi-camera rig. Upper HD pilot camera on its own pan-tilt unit. Lower HD pilot camera and a 4K pilot camera on the same pan-tilt unit. A stereo-camera housing.]{\includegraphics[ width =0.4\columnwidth]{figures/ROV.png}} } &
%\subfloat[Prototype of multi-camera rig]{\includegraphics[ width = 0.45\columnwidth]{figures/ROV_rig_zoom.png}} \\

%& \subfloat[ROV Control Room. Photograph by Ole Martin Wold]{\includegraphics[width = 0.45\columnwidth]{figures/rov_control_rom.jpeg}}
%\end{tabular}
%\caption{}
%\label{fig:system_overview}
%\end{figure}

In particular, visual imaging systems have been instrumental in enhancing spatial perception, facilitating simultaneous localization and mapping (SLAM) strategies that build rich world models while accurately estimating the vehicle’s state. When incorporated into ROV and AUV platforms, vision-based navigation and mapping not only supports operator decision-making but also unlocks downstream applications—such as online planning, virtual environment generation, and semantic scene understanding. These functionalities have further implications for broader marine research endeavors, including habitat monitoring, environmental sampling, and large-scale ocean ecosystem analysis \cite{habitat_mapping}\cite{sample_collection_with_3d_vision}.

Despite the progress made, there remain substantial hurdles to effective vision-based mapping in underwater contexts. Light attenuation, backscatter, low-light conditions, and feature-poor regions impose significant visual degradation \cite{mobley_ocean_optics} that challenges existing algorithms. Meanwhile, issues like camera synchronization, unknown or variable camera parameters (e.g., zoom, pan, and tilt), and the hardware complexity of multi-camera setups exacerbate these difficulties. Many current visual mapping systems have been designed primarily for monocular or stereo cameras, limiting their scalability and effectiveness when multiple cameras are installed on a single vehicle.

To address these challenges, we explore multi-modal sensing by integrating data from diverse sensors and incorporating higher-level semantics to improve both spatial perception and overall situational awareness. In this paper, we introduce underwater datasets collected during multiple field campaigns and dockside testing using NTNU’s work-class ROV, which is equipped with vision, inertial, and acoustic sensors. We then showcase a range of visual mapping methodologies applied to these datasets, detailing the sensor configurations, the acquired data, and the resulting dense 3D maps. These findings underscore the significant role of real-time mapping in enhancing marine exploration and situational awareness, ultimately paving the way for more robust long-term underwater autonomy.

% \begin{figure*}[h!]
%     \centering
%      \begin{tabular}{c c}
%       % First column: large image spanning two rows
%       \multirow{2}{*}{\begin{minipage}[t]{0.4\textwidth}
%         \includegraphics[width=\textwidth]{figures/ROV.png}
%         \captionof{figure}{Minerva II}
%         \label{fig:minerva2}
%       \end{minipage}}
%       &
%       % Second column (top)
%       \begin{minipage}[t]{0.5\textwidth}
%         \includegraphics[width=\textwidth]{figures/ROV_rig_zoom.png}
%         \captionof{figure}{Prototype of Multi-Camera Rig}
%         \label{fig:multi_camera_rig}
%       \end{minipage}
%       \\
%       % Second column (bottom)
%       &
%       \begin{minipage}[t]{0.5\textwidth}
%         \includegraphics[width=\textwidth]{figures/rov_control_rom.jpeg}
%         \captionof{figure}{ROV Control Station. Photographer Ole Martin Wold.}
%         \label{fig:rov_control_station}
%       \end{minipage}
%     \end{tabular}
%     \caption{Caption}
%     \label{fig:system_overview}
% \end{figure*}

\section{System Overview}
\subsection{ROV Platform}

Minerva II is a 2.5-tonne work-class ROV operated by the Applied Underwater Robotics Laboratory at NTNU, it can be seen in figure \ref{fig:minerva2}. It's a modified Sperre 100K-SubFigther and carries two manipulator arms in the front, a Schilling ORIN7P on the starboarrd side, and a Schilling RigMaster on the port side. In terms of sensor equipment, the ROV is outfitted with three forward-looking cameras mounted on pan-tilt units, a Norbit WBMS forward-looking sonar, a Teledyne Workhorse Navigator Doppler Velocity Log (DVL), a Valeport VA500 pressure sensor, and a Safran STIM300 Inertial Measurement Unit (IMU). 

The ROV is controlled from a control room which is depicted in figure \ref{fig:minerva2_control_room}. It can be controlled in 4-DOF, and will only have small deviations from its neutral position in roll and pitch. 

\subsection{Multi-camera Prototype}

Although the ROV is equipped with three forward-looking cameras, they are all mounted on pan-tilt units that are actively utilized during deployments. Consequently to reduce complexity of the multi-camera problem, a prototype of a rigid multi-camera rig was developed and employed to collect datasets during field deployments. This rig is depicted in figure \ref{fig:multi_camera_rig}.

The rig comprises three Deep-Water Explorer (DWE) StellarHD cameras, which can be mounted with downward tilts of 0, 30, 45 and 60 degrees relative to the horizontal plane. The outer cameras can also be tilted outward to expand the system's total field of view. The DWE cameras provide leader-follower frame synchronization. By designating one camera as the leader and the others as followers, and connecting their trigger cables, the cameras synchronize their frames automatically.

The cameras are connected to a Blue Robotics enclosure functioning as a junction box, which houses a Khadas VIM4 single-board computer and a STIM300. Both the cameras and the STIM300 are interfaced with the Khadas. This module is integrated with the rest of the ROV system via a 1 Gbit/s Ethernet connection. During deployments, the module was mounted atop the ROV, as illustrated in Figure \ref{fig:minerva2}.

The camera rig was calibrated over two sessions using a large checkerboard and the Kalibr toolbox \cite{kalibr}. The initial session was conducted in air to determine the extrinsic parameters between the fixed cameras. Subsequently, a calibration was performed in a saltwater pool to find the intrinsic parameters for each individual camera.

\subsection{Data Collection and Synchronization}

The computers utilized for sensors and data collection in this study were all synchronized to the same Network Time Protocol (NTP) server. All used sensors had their own ROS driver, and during data collection data was recorded directly to a ROS bag. Every message were stored with a time-stamp. Camera streams were transmitted topside prior to initializing data collection, enabling the testing of various lighting conditions to achieve optimal quality before starting to record datasets. The collected datasets are described in the next section.

\begin{figure*}[t]
\centering

\begin{tabular}{c}
\subfloat[]{\includegraphics[width=0.95\linewidth]{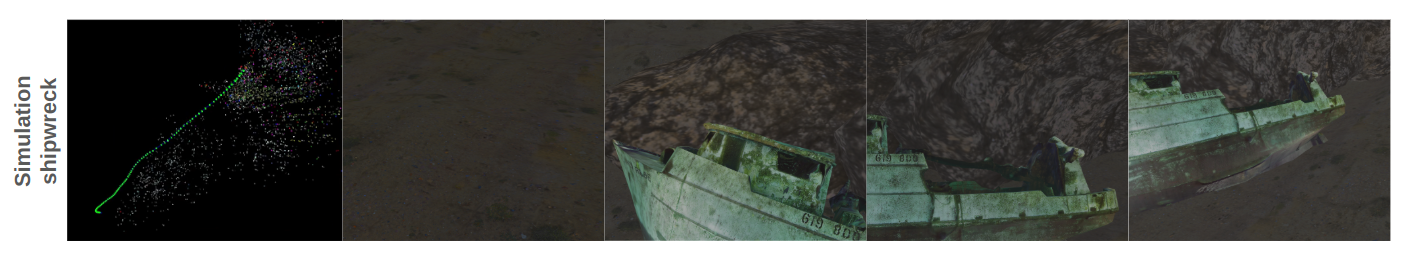}} \\
\subfloat[]{\includegraphics[width=0.95\linewidth]{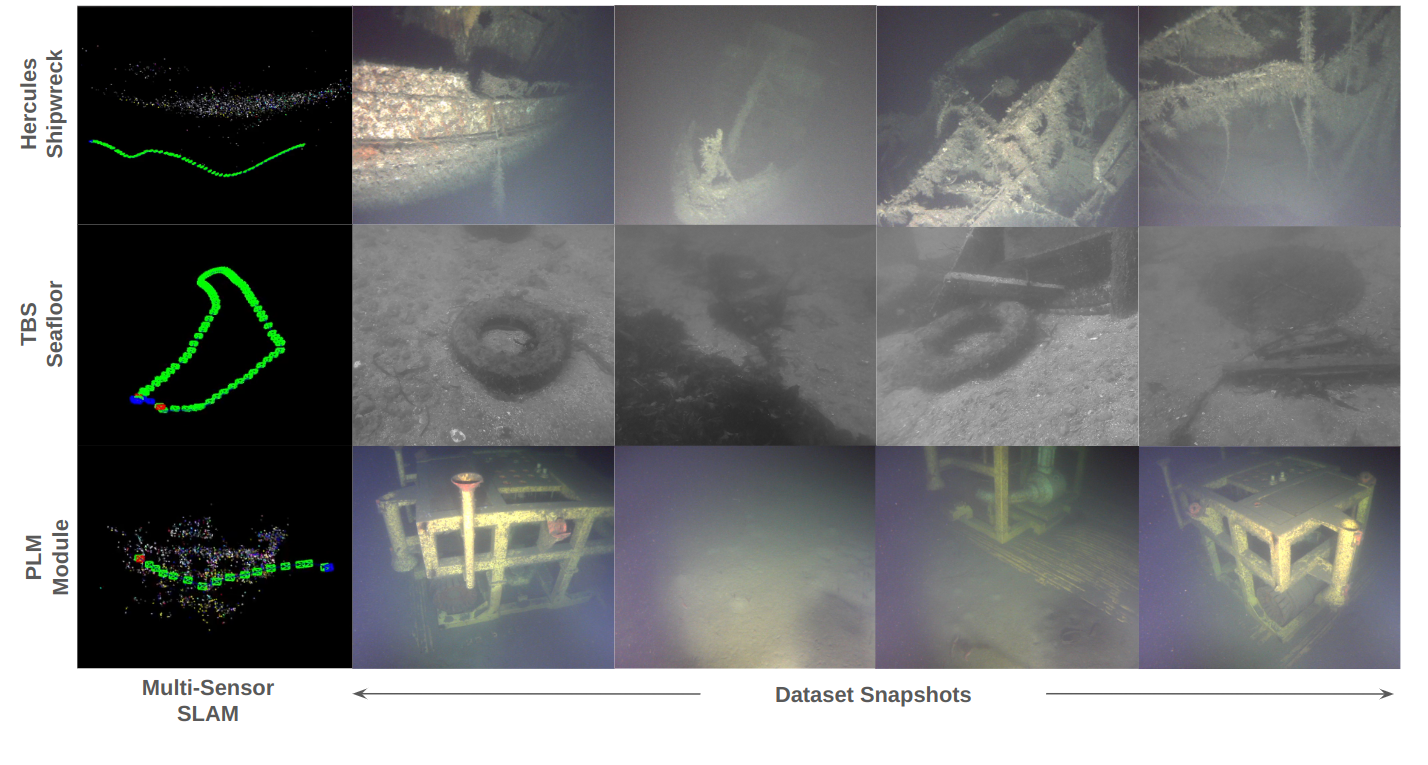}}
\end{tabular}
\caption{Snapshots from both the synthetic (a) and real-world field datasets (b), alongside the multi-camera, multi-sensor (inertial and DVL) SLAM results. The first column shows the estimated trajectory and the sparse point cloud generated during SLAM, while the remaining columns depict various scenes captured during data collection, highlighting challenges such as uneven and low lighting, limited visibility, haze, and backscatter.}
\label{fig:data_results}

\end{figure*}
\begin{figure*}[]
\centering
\includegraphics[width=0.95\linewidth]{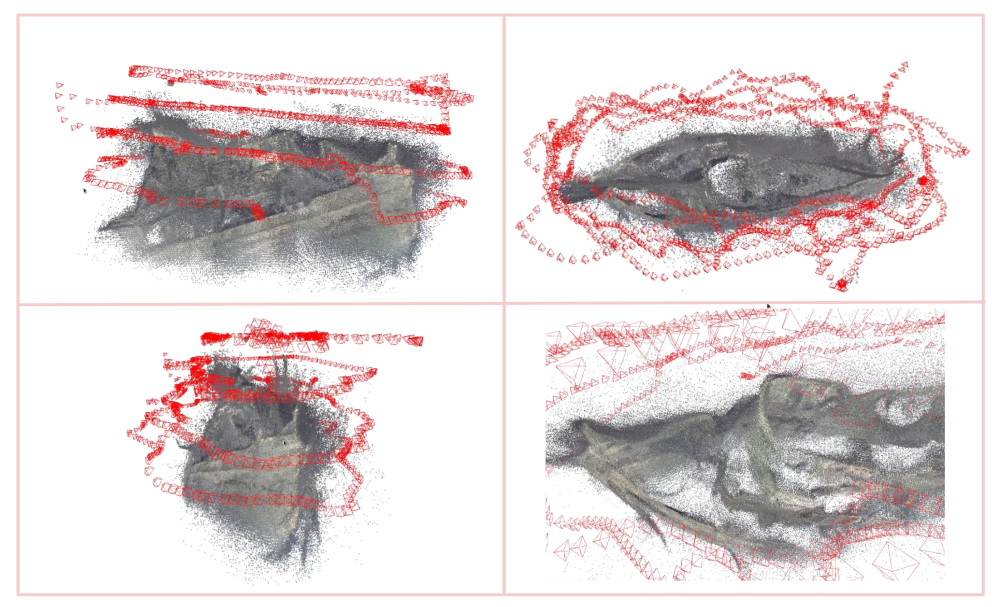}

\caption{The final dense 3D reconstruction of the Hercules Shipwreck captured from various viewing angles. The red triangles represent the camera poses, and we can see the concentric circular trajectories the ROV took around the shipwreck.}
\label{fig:data_results_droid}

\end{figure*}

\begin{table}[t!]
% \captionsetup{font={footnotesize}}
\caption{Description of various sensors and their settings used to collect our datasets. Note that OAK-D sensor is available only in the TBS Seafloor dataset.}
\label{tab:sensor_setup}
\resizebox{\columnwidth}{!}{%
\begin{tabular}{llll}
\hline
\textbf{Sensor} &
  \textbf{No} &
  \textbf{Type} &
  \textbf{Description} \\ \hline \hline
Camera &
  3 &
  \begin{tabular}[c]{@{}l@{}}DWE \\ Stellar  UVC\end{tabular} &
  \begin{tabular}[c]{@{}l@{}}2 MP color cameras \\ with a resolution of 1600 x 1200 \\ and FoV of 82 $^\circ$ (in water) at 30 hz.\end{tabular} \\
  IMU &  1 &  \begin{tabular}[c]{@{}l@{}}STIM \\ 300\end{tabular} &  9-DOF IMU running at 500 Hz. \\ 
 DVL &   1 &  \begin{tabular}[c]{@{}l@{}}Workhorse\\ Navigator DVL \end{tabular} &
  \begin{tabular}[c]{@{}l@{}}DVL running at 7 Hz max \end{tabular} \\
\begin{tabular}[c]{@{}l@{}}Oak-D\\ camera\end{tabular} &
  1 & Stereo &
  \begin{tabular}[c]{@{}l@{}} 1 MP cameras with resolution of \\ 1280 x 800 at 60 Hz, \\ BNO085 9-DOF IMU \end{tabular} \\ \hline

\end{tabular}%
}

\end{table}
% Please add the following required packages to your document preamble:
% \usepackage{multirow}
\begin{table}[]
\caption{Description of various attributes of our collected datasets including camera setups, location, date and duration.}
\label{tab:dataset}
\resizebox{\columnwidth}{!}{
\begin{tabular}{|l|l|l|l|l|}
\hline
\multicolumn{1}{|c|}{\textbf{Label}} &
  \multicolumn{1}{c|}{\textbf{Date}} &
  \multicolumn{1}{c|}{\textbf{Sensor setup}} &
  \multicolumn{1}{c|}{\textbf{\begin{tabular}[c]{@{}c@{}}Duration\\ (sec)\end{tabular}}} &
  \multicolumn{1}{c|}{\textbf{Loops}} \\ \hline
Synthetic wreck &
  NA &
  Stereo camera setup &
  71 &
  No \\ \hline
\multirow{2}{*}{Hercules Shipwreck} &
  18 June 2024 &
  \begin{tabular}[c]{@{}l@{}}3 cams pitched 30 deg. \\ Center facing forward,\\  left and right facing\\  away with 30 deg yaw\end{tabular} &
  285 &
  Yes \\ \cline{2-5} 
 &
  21 June 2024 &
  \begin{tabular}[c]{@{}l@{}}Center and right camera \\ both facing forward like \\ stereo pair. Both pitched \\ forward by 45 deg.\end{tabular} &
  4260 &
  Yes \\ \hline
PLM Module &
  21 june 2024 &
  \begin{tabular}[c]{@{}l@{}}Center camera facing \\ forward, right camera\\  facing away from center\\  by 30 deg. Both pitched \\ forward by 45 deg.\end{tabular} &
  1072 &
  Yes \\ \hline
TBS Seafloor &
  19 April 2024 &
  \begin{tabular}[c]{@{}l@{}}Stereo cameras\\ facing forward (Oak-D)\end{tabular} &
  537 &
  Yes \\ \hline
Pipeline &
  27 June 2024 &
  \begin{tabular}[c]{@{}l@{}}3 cams pitched 45 deg. \\ Center facing forward,\\  left and right facing\\  also facing forward.\end{tabular} &
  736 & 
  No \\ \hline
\end{tabular}
}

\end{table}

\section{Field campaigns and datasets}

Our dataset collection comprises both real-world data gathered during field campaigns and synthetic data generated in a Gazebo simulation environment. \Cref{tab:sensor_setup} gives description of the sensors used for data collection and \Cref{tab:dataset} provides a high-level summary of these datasets.

\subsection{Real-world data} 
We present three primary real-world datasets collected from onboard sensors on an ROV. These include surveys of a shipwreck, a subsea infrastructure component known as the Pig Loop Module (PLM), a short deck test near the Trondheim Biological Station (TBS) at NTNU, and a survey of an underwater pipeline.

\textbf{Hercules Shipwreck} This dataset consists of visual imagery, IMU, and DVL data captured while driving the ROV around the shipwreck in concentric circles with varying altitude. Care was taken to ensure full vertical coverage of the wreck, and consecutive circles were designed to have overlapping images that facilitate image registration. The survey lasted approximately 71 minutes, providing extensive data for testing mapping and localization algorithms in a visually diverse underwater environment.

\textbf{PLM module Survey } This dataset was collected using a similar circular-survey strategy, although limited to two concentric loops due to the smaller size of the infrastructure module. After the main survey, the ROV was piloted away from the PLM, capturing largely featureless imagery of the seafloor and water column under low-light conditions. It then returned to the PLM for another loop, creating a challenging sequence that blends feature-rich segments around the module with feature-poor intervals. This makes the dataset particularly valuable for evaluating 3D perception, SLAM, rendering, and sensor-fusion algorithms under varying conditions.

\textbf{TBS Seafloor} Gathered off the dock at TBS, the TBS dataset was recorded using only OAK-D stereo cameras  The ROV followed a short loop close to the seafloor, capturing images under relatively bright sunlight but encountering turbid water that introduced haziness and degraded image quality at greater distances. This dataset highlights the difficulties posed by variable optical conditions in shallow-water environments.

\textbf{Pipeline data} The dataset was gathered off the dock at TBS, following an underwater pipeline from its outlet to the shore. It was recorded using the multi-camera rig as the ROV tracked the pipeline in a northwest direction, ascending from a depth of 70 meters to 30 meters. This dataset is valuable for testing various semantic extraction methods to identify pipelines and pipeline supports in seabed imagery. 

\subsection{Simulation Data}
This data is generated using ROS gazebo, an open source robotics simulator. We used a simulated model of the ROV with onboard sensors and ROS data stack . This ROV is operated in a virtual environment containing a shipwreck on the seafloor as shown in the figure. As the ROV is moving and exploring the environment, data from the simulated onboard sensors was recorded into a rosbag. Although not a substitute for real-world missions, this simulated environment provides a cost-effective platform for testing and refining algorithms before conducting field operations, thereby reducing the risk and expense of on-site deployments \cite{from_simulated_waters}.

\section{Vision based Perception}
Our primary objective is to enhance the situational awareness of autonomous underwater systems by integrating multi-modal sensing and advanced learning-based techniques to enable real-time, robust perception and navigation in complex environments under context-aware constraints. We primarily focus on vision based perception, but augment it with multiple other sensing modalities to mitigate the visual degradation faced by robots in underwater environments. Our situational awareness system must provide real-time performance, resilience to sensor degradation and/or failures, higher-level semantic understanding, and 3D spatial mapping—all essential for accurate robot state estimation and comprehensive environmental awareness.

In this section, we present preliminary results from vision-based SLAM in underwater settings. The methods we present encompass both geometric and learning-based approaches, demonstrating the feasibility of fusing multi-camera and multi-sensor data for improved performance in challenging marine conditions.

\subsection{Multi-camera multi-sensor mapping }
Visual SLAM has become a mature research field, producing numerous state-of-the-art techniques that target monocular, stereo, and RGB-D configurations\cite{he2020review}\cite{campos2021orb}. More recently, interest has grown in multi-camera SLAM systems\cite{kuo2020redesigning}\cite{urban2016multicol} due to their ability to circumvent single-point failures due to degenerate motion, variations in illumination, feature-scarce environments, and the presence of dynamic objects. Indeed, multiple studies have shown that multi-camera sensing can enhance SLAM performance in terrestrial datasets collected in urban environments. Underwater platforms can get significant benefits from multi-camera SLAM, as it helps mitigate the visibility challenges inherent to submerged operations.  

We explored one of the latest multi-camera SLAM frameworks, MCSLAM\cite{kaveti2023design}, which uses a generalized camera model to represent an arbitrary multi-camera setup as a collection of unconstrained rays, thus avoiding assumptions about a particular geometry. Building on MCSLAM, we extended its capabilities by incorporating additional sensing modalities such as IMU and DVL data. The MCSLAM framework consists of two main components:
\begin{itemize}
    \item \textbf{Front-End:} Handles initialization, feature extraction, and matching, producing initial pose and landmark estimates for each frame.
    \item \textbf{Back-End:} Constructs a factor graph and performs non-linear optimization on the initial estimates to produce refined, accurate solutions.
\end{itemize}

We retained the original MCSLAM front-end while augmenting the back-end with custom IMU and DVL factors. By tightly integrating these sensors, our system not only benefits from improved resilience under severe visual degradation but also leverages complementary data sources to bolster SLAM performance. This comprehensive, multi-sensor approach offers a more robust and context-aware solution for real-time underwater navigation and 3D perception.

\Cref{fig:data_results} illustrates the outcomes of the multicamera, multisensor SLAM system across various datasets, displaying both the estimated trajectory and a sparse point cloud of the reconstructed surfaces. Acquiring ground-truth data in real-world field experiments poses substantial challenges, so we primarily rely on the continuity of tracking and the structure of the reconstructed point clouds as qualitative performance indicators. In the PLM Module and Hercules shipwreck datasets, the distinctive shapes of the ship’s hull and the module’s gridlike structure are clearly visible in the generated point clouds. The TBS seafloor dataset does not include a corresponding sparse point cloud; however, the ROV’s trajectory is deliberately designed to return to its starting position, facilitating a loop closure event. This loop closure is highlighted by the appearance of blue keyframes in the estimated trajectory, confirming the system’s ability to robustly track and match visual features.

\begin{figure}
\centering
\includegraphics[width=\columnwidth]{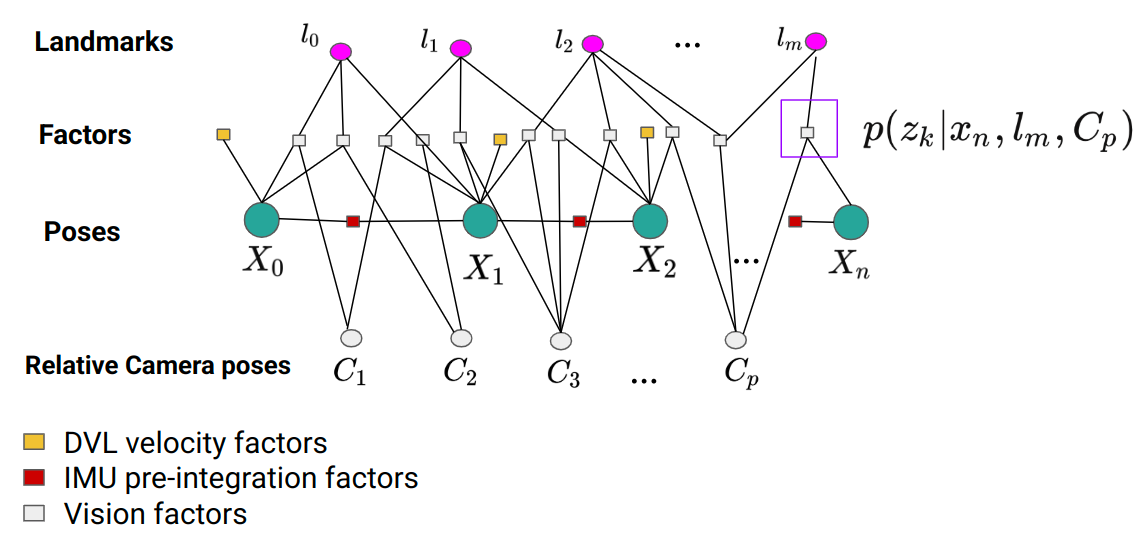}
\caption{Factor graph of the multi-camera back-end with IMU and DVL factors. The poses $X_i$, landmarks $l_j$ and the relative camera poses $C_p$ are the variables to be optimized.}
\label{fig:backend}

\end{figure}

\subsection{Learning based Optical Flow}
In cases of extreme visual degradation, both classical feature-based methods and conventional optical flow techniques often fail. Recent advances in deep learning, however, have enabled a host of data-driven optical flow approaches (e.g., Flownet\cite{dosovitskiy2015flownet}, Gmflow\cite{xu2022gmflow},Raft\cite{teed2020raft}) that learn complex representations from training data. These learned models tend to be more robust to noise and less bound by classical assumptions such as small displacements and strict brightness constancy. Moreover, they are more generalizable and adaptable to variations in lighting, texture, and motion—conditions that frequently occur in challenging real-world underwater environments.

Motivated by these advantages, researchers have begun integrating deep optical flow backbones into visual odometry pipelines. One such approach, DROID-SLAM \cite{droidslam}, has achieved state-of-the-art results on numerous benchmark datasets. A defining feature of DROID-SLAM is its dense bundle adjustment layer, which employs GPU-based optimization to closely mirror the principles of classical SLAM methods.

We evaluated DROID-SLAM on our underwater datasets. As shown in \Cref{fig:data_results_droid}, its vision-only pipeline produces a high-quality, dense 3D reconstruction of the Hercules shipwreck, including clear circular trajectories of the ROV around the ship’s hull. Although these results are promising, the vanilla DROID-SLAM framework has certain limitations. First, the final reconstruction is generated only after the system’s online bundle adjustment stage; prior to global refinement, each circular pass exhibits noticeable drift due to visual degradation. Second, DROID-SLAM does not inherently support loop closures or relocalization—both standard features in many classical SLAM solutions. As a result, if the ROV temporarily leaves the visible surface (for instance, traversing an unfeatured water column) and returns later, the algorithm may fail to establish correspondences with the existing map. This shortcoming is particularly pronounced in the PLM dataset, where the ROV intermittently departs from the module and then revisits it, causing DROID-SLAM to lose track.

% \begin{figure}[t]
% \centering
% \includegraphics[width=\columnwidth]{figures/learning_EKF.png}

% \caption{}
% \label{fig:data_results_droid}

% \end{figure}

\subsection{Semantic annotations}

Real-time, robust spatial perception of the underwater environment is essential for enhancing navigation and is crucial for complex tasks such as collision avoidance and path planning for AUVs. However, to accomplish even more intricate tasks, such as infrastructure inspection and valve intervention, AUVs must not only understand the spatial aspects of the environment but also recognize and interpret the objects within it. 

Semantic segmentation is a mature technique, with modern approaches relying on deep-learning, for extracting semantic information from images. Current state-of-the-art segmentation methods are employed in autonomous vehicles, frequently in conjunction with LiDAR technology \cite{camera_and_lidar_1}\cite{camera_and_lidar_2}\cite{camera_and_lidat_3}, to develop a comprehensive understanding of their operational environment. This approach not only facilitates the spatial mapping of their surroundings but also enables the differentiation between static and different dynamic objects within the scene. 

\begin{figure}[h]
    \centering
    \begin{minipage}{0.25\textwidth}
        \centering
        \includegraphics[width=\textwidth]{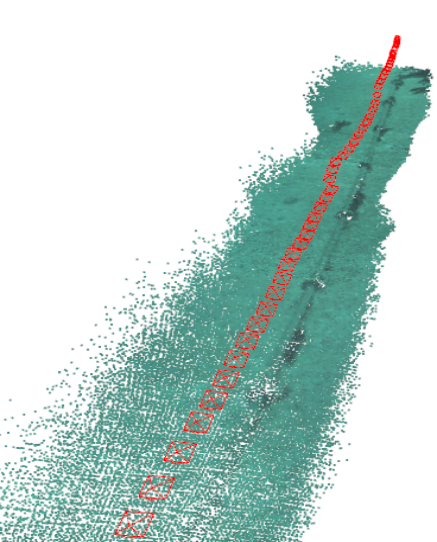} 

    \end{minipage}\hfill
    \begin{minipage}{0.25\textwidth}
        \centering
        \includegraphics[width=\textwidth]{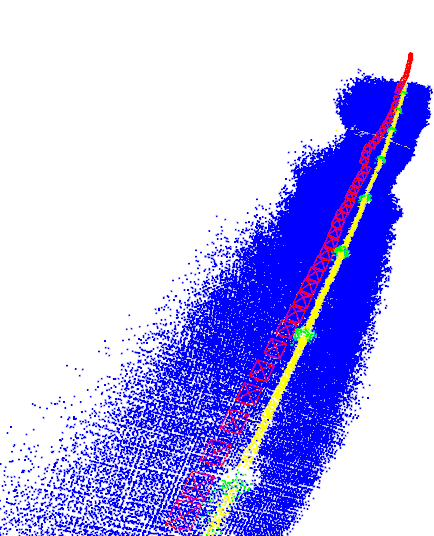}
    \end{minipage}
    
    \caption{Point cloud generated by DROID-SLAM with and without semantic information projected onto the cloud. The pipeline points is set to yellow, the pipeline supports to green, and the seabed to blue.}
    \label{fig:semantic}
\end{figure}

The ability to robustly generate real-time semantic representations in underwater environments in a similar manner is a crucial step toward enhancing the autonomy of underwater vehicles. As an initial naive approach to showcase a semantic 3D representation, we employed the pipeline dataset. We trained a DeepLavx3 model \cite{deeplab} to classify each pixel in an image as either pipeline, pipeline support or background. By running this model in parallel with DROID-SLAM, we could project the pixel classes onto the point cloud by using DROID-SLAMs estimated camera poses and depth maps. The resulting semantic point cloud is illustrated in figure \ref{fig:semantic}.

\section{Discussion} 
This section summarizes our findings from three distinct, yet interrelated, perception approaches: (1) a multi-camera, multi-sensor SLAM framework based on classical geometric methods, (2) a learning-based visual odometry (VO) pipeline (DROID-SLAM), and (3) a semantic mapping procedure that projects pixel-level semantic labels onto a 3D reconstruction. By comparing these methods and analyzing their respective strengths and weaknesses, we derive insights into how future underwater perception systems can be improved.

\subsection{Learning-Based vs. Geometric Approaches}
Our classical multi-camera, multi-sensor SLAM framework employs a sparse feature-matching front end combined with factor-graph optimization and loop-closure capabilities. This geometric method offers several key advantages in underwater scenarios. It relies on direct feature extraction without extensive training or network re-tuning to adapt to new domains. It minimizes drift through recognizing previously mapped areas via loop closures. Finally, providing robust performance by fusing data from multiple cameras and sensors.
Nevertheless, geometric methods can struggle in low-visibility conditions common to underwater operations. When the scene becomes too featureless or overly obscured by backscatter and turbidity, sparse feature matching becomes brittle, leading to lost tracks or poor map quality.
By contrast, our learning-based VO approach (DROID-SLAM) relies on learned optical flow, thus improving robustness under visually challenging conditions. However, the approach lacks loop-closure capabilities and supports only a single-camera input, causing it to accumulate drift over longer sequences. Without an explicit mechanism to fuse inertial or acoustic data, errors in pose estimation can compound quickly when external visual data is lost for extended periods
One promising direction is integrating learning-based feature extraction with geometric SLAM back ends to leverage both data-driven robustness and the established loop-closure and sensor-fusion mechanisms of classical pipelines. Such a hybrid system could offer greater resilience to visual degradation while maintaining long-term accuracy via loop closure.

\subsection{Calibration Challenges in Multi-Camera Systems}
An additional concern, particularly relevant to multi-camera setups, is camera calibration. When cameras are shared between operational teams—who may change zoom, pan, and tilt settings on the fly—intrinsic and extrinsic parameters can shift, breaking many traditional SLAM assumptions. Underwater environments compound this issue further, due to refraction and varying camera enclosures.
For a reliable spatial perception system, it is crucial to handle these dynamic calibration issues in real time. Potential solutions include online auto-recalibration that updates camera parameters when changes are detected, and adaptable sensor fusion to seamlessly incorporate evolving camera parameters without requiring full re-initialization.

\subsection{Underwater semantic segmentation}

While the resulting semantic point cloud in figure \ref{fig:semantic} looks fairly good, the pipeline dataset provides very good visibility and lighting conditions. In addition the segmentation model is trained of nearby similar pipelines. 

However, under harsher conditions, such as poor lighting or reduced visibility, the image quality may degrade significantly, potentially leading to failures in the segmentation network. Some of this can be mitigated with proper image pre-processing, but there are situations, for example if the thrusters stir up significant amounts dust and particles from the seabed, where no amount of image processing would help. These cases necessitates the development of a robust mechanism to identify and discard low-quality data that could compromise the accuracy of the segmentation. 

Furthermore, the scarcity of underwater training data poses a significant challenge, as it may limit the network's ability to generalize effectively. Exploring the generation of synthetic data to augment existing datasets, as well as leveraging transfer learning techniques to adapt models trained on larger, more diverse datasets, will be crucial research directions.

Since semantic segmentation performs segmentation on images individually, and don't rely on consistent visibility like SLAM, it can potentially be utilized to aid SLAM navigation. When visibility deteriorates, the SLAM algorithm may lose its tracking, leading to disorientation. Upon the return of clear conditions, the camera might no longer recognize previously seen features, causing the SLAM system to lose its positional awareness. However, by leveraging semantic segmentation, an autonomous agent can identify nearby objects and deduce that if it has observed any of those objects earlier. This recognition allows the agent to navigate towards the previously seen objects, aiding in the rediscovery of its previous map and effectively closing the loop, thereby enhancing the robustness of underwater navigation in dynamic visibility conditions.

\section{Conclusion}
This paper presents the datasets collected during field campaigns in the Trondheim Fjord aboard the Gunnerus Research vessel using the work-class ROV Minerva. Through benchmarking various vision-based SLAM approaches on these datasets, we highlighted both the successes and the remaining limitations in current underwater perception methods. Drawing on our experimental results and hands-on field experience, we also identified pressing research challenges that must be addressed to achieve robust perception and situational awareness in underwater environments. Overcoming these challenges will be crucial to enabling long-term autonomous operations and advancing the capabilities of future underwater robotic systems.

%%%%%%%%%%%%%%%%%%%%%%%%%%%%%%%%%%%%%%%%%%%%%%%%%%%%%%%%%%%%%%%%%%%%%%%%%%%%%%%%

%\addtolength{\textheight}{-12cm}   % This command serves to balance the column lengths
                                  % on the last page of the document manually. It shortens
                                  % the textheight of the last page by a suitable amount.
                                  % This command does not take effect until the next page
                                  % so it should come on the page before the last. Make
                                  % sure that you do not shorten the textheight too much.

%%%%%%%%%%%%%%%%%%%%%%%%%%%%%%%%%%%%%%%%%%%%%%%%%%%%%%%%%%%%%%%%%%%%%%%%%%%%%%%%

%%%%%%%%%%%%%%%%%%%%%%%%%%%%%%%%%%%%%%%%%%%%%%%%%%%%%%%%%%%%%%%%%%%%%%%%%%%%%%%%

%%%%%%%%%%%%%%%%%%%%%%%%%%%%%%%%%%%%%%%%%%%%%%%%%%%%%%%%%%%%%%%%%%%%%%%%%%%%%%%%
%\section*{APPENDIX}

%%%%%%%%%%%%%%%%%%%%%%%%%%%%%%%%%%%%%%%%%%%%%%%%%%%%%%%%%%%%%%%%%%%%%%%%%%%%%%%%
	
\bibliographystyle{IEEEtran}
\bibliography{references}

\end{document}